\documentclass[10pt,a4paper]{article}

\usepackage[]{lrec-coling2024} % this is the new style

\usepackage{nameref}
\usepackage{epstopdf}
\usepackage[utf8]{inputenc}
\usepackage{xstring}
\usepackage{color}
\usepackage{microtype}
\usepackage{inconsolata}
\usepackage{amsmath}
\usepackage{cleveref}
\usepackage{arydshln}
\usepackage{booktabs}
\usepackage{comment}

\usepackage{multibib}
\newcites{languageresource}{Language Resources}
\usepackage{graphicx}
\usepackage{subcaption}
\usepackage{tabularx}
\usepackage{soul}
\usepackage[]{titlesec}
\title{DADIT: A Dataset for Demographic Classification of Italian Twitter Users and a Comparison of Prediction Methods}

\name{Lorenzo Lupo, Paul Bose, Mahyar Habibi, Dirk Hovy, Carlo Schwarz} 

\address{All authors are based at Bocconi University \\
         Via Roberto Sarfatti 25, 20136 Milano, Italy \\
         \{paul.bose, mahyar.habibi, lorenzo.lupo, dirk.hovy, carlo.schwarz\}@unibocconi.it\\}

\abstract{
    Social scientists increasingly use demographically stratified social media data to study the attitudes, beliefs, and behavior of the general public. To facilitate such analyses, we construct, validate, and \href{https://github.com/lorelupo/twitter_user_classification}{release publicly} the representative DADIT dataset of 30M tweets of 20k Italian Twitter users, along with their bios and profile pictures. We enrich the user data with high-quality labels for gender, age, and location. DADIT enables us to train and compare the performance of various state-of-the-art models for the prediction of the gender and age of social media users. In particular, we investigate if tweets contain valuable information for the task, since popular classifiers like M3 don't leverage them. Our best XLM-based classifier improves upon the commonly used competitor M3 by up to 53\% F1. Especially for age prediction, classifiers profit from including tweets as features. We also confirm these findings on a German test set. \\ 
    \newline \Keywords{Twitter data, demographic prediction, language models, multimodal classification}
}

\begin{document}
\maketitleabstract

%%%%%%%%%%%%%%%%%%%%%%%%%%%%%%%%%%%%%%%%%%%%%%%%%%%%%%%%%%%%%%%%%%%%%%%%%%%%%%%%%%%%%%%%%%%%%%
%%%%%%%%%%%%%%%%%%%%%%%%%%%%%%%%%%%%%%%%%%%%%%%%%%%%%%%%%%%%%%%%%%%%%%%%%%%%%%%%%%%%%%%%%%%%%%

%%%%%%%%%%%%%%%%%%%%%%%%%%%%%%%%%%%%%%%%%%%%%%%%%%%%%%%%%%%%%%%%%%%%%%%%%%%%%%%%%%%%%%%%%%%%%%
%%%%%%%%%%%%%%%%%%%%%%%%%%%%%%%%%%%%%%%%%%%%%%%%%%%%%%%%%%%%%%%%%%%%%%%%%%%%%%%%%%%%%%%%%%%%%%
\section{Introduction \label{sec:intro}}
%%%%%%%%%%%%%%%%%%%%%%%%%%%%%%%%%%%%%%%%%%%%%%%%%%%%%%%%%%%%%%%%%%%%%%%%%%%%%%
% European societies have been strained by discontent with democracy, mistrust in institutions, populist parties, and rising inequality for a decade. The Covid-19 pandemic only worsened many of these issues as governments' anti-Covid-19 measures disparately impacted different societal groups. Social scientists increasingly make use of social media data to study these issues. 
Social media data are a valuable support for economic and social science research, and they have been leveraged extensively in the literature \cite[see][for a review]{steinert2018twitter}. For many use cases, social media users need to be profiled along their demographics to study differences within and between groups \cite[e.g.,][]{holmberg2015gender,usher2018twitter,allen2022birds}. For example, studying social and economic inequalities in a country requires stratifying users across gender, age, and geographic location.

Social media users' demographics are typically inferred from the user's profile information, such as the username \cite{lockhart2023name}, their posted pictures \citep{you2014images}, or their network \citep{al2012homophily,barbera2015birds}. 
One of the most popular tools, the M3 classifier \cite{wang2019demographic}, predicts age and gender multimodally, leveraging usernames, users' bios, and pictures. 
However, these tools often depend on the availability of biographical information or profile images that contain identifiable pictures of the actual users. Whenever this information is absent, the accuracy of these approaches drops significantly, or they become unusable altogether. For example, almost 27\% of US adults do not provide any text in their Twitter profile, and 14\% do not provide any photo (e.g., rear shots or nature photos; see \citet{pew_2022}); a lot of information is also missing on Italian profiles (\Cref{tab:data}). Both in the absence and the presence of these features, users' posts provide valuable information for user characteristics, but surprisingly, little attention has been paid to leveraging this information.

We bridge this gap by collecting and releasing DADIT: a GDPR-compliant database of around 30M tweets from 20K representative Italian Twitter users, along with their profile pictures and bios.
% Our dataset distills a broader initial collection of 830.9M tweets covering 2.1M users, available upon request, on which we ran heuristics to extract gender, age, and approximate location of the users to allow stratification along demographic lines.
DADIT contains high-quality, manually evaluated labels for gender and age, along with location labels for 43\% of the users. Given the recent challenges of gathering Twitter data for academic research, our collection supports future research. 

Leveraging this data, we systematically compare gender and age classification methods.
We are particularly interested if and by how much the inclusion of users' posts (tweets) improves the prediction of gender and age by comparing modern text classification models \citep{brown2020language,barbieri2021xlm,chung2022scaling} to traditional
users' classifiers: M3 \citep{lockhart2023name} and another popular computer vision (CV) classifier \citep{levi2015age}.
We find that models leveraging bios and tweets often outperform models based on bios and pictures. In particular, a text classifier finetuned on our Italian dataset improves upon the state-of-the-art M3 by a significant margin. This finding underscores the usefulness of text-abundant, demographically annotated datasets like DADIT.
Finally, we demonstrate that ensembling multiple models' predictions further improves classification performance.
To strengthen the generalizability of our results, we also experiment on a German test set, corroborating the Italian results.

Our contributions are:
\begin{enumerate}
    \item We release DADIT, an extensive database of around 30M tweets from 20K Italian Twitter users, including profile photos, bios, and tweets, enriched with high-quality demographic information (age, gender) and geographic information (Italian regions and, in some cases, municipalities). The data will be released in GDPR and Twitter-compliant anonymized format (see also Section \ref{sec:ethics}).
    \item We systematically compare user attribute prediction methods in Italian and German. We demonstrate that including tweet texts brings significant improvements over profile-based classification methods like M3, especially when training data like DADIT are available.
\end{enumerate}

%%%%%%%%%%%%%%%%%%%%%%%%%%%%%%%%%%%%%%%%%%%%%%%%%%%%%%%%%%%%%%%%%%%%%%%%%%%%%%%%%%%%%%%%%%%%%%
%%%%%%%%%%%%%%%%%%%%%%%%%%%%%%%%%%%%%%%%%%%%%%%%%%%%%%%%%%%%%%%%%%%%%%%%%%%%%%%%%%%%%%%%%%%%%%%%%%%%%%%%%%%%%%%%%%%%%%%%%%%%%%%%%%%%%%%%%%%%%%%%%%%%%%%%%%%%%%%%%%%%%%%%%%%%
\section{The Dataset} 
\label{sec:data}

%%%%%%%%%%%%%%%%%%
\subsection{Collection}\label{sec:data:collection}
%%%%%%%%%%%%%%%%%%
We collected the data through the Twitter API in early 2023. We used the Twitter accounts of 802 Italian politicians and parties as a source list for user IDs.\footnote{See \Cref{sec:appendix_data} for an overview of the number of accounts and the total number of followers by party.} 
For each account, we collected the user IDs of followers.
In total, we obtained 3.2M unique user IDs. We then collected each user's profile and tweets as a second step. Note that we did not obtain tweets for all users for one of three reasons: 1) some users do not tweet, 2) some accounts were deleted before we were able to scrape them, or 3) the Twitter APIs were shut down in early 2023 as a result of Elon Musk's takeover of Twitter which prevented further data collection. Nonetheless, we obtained around 830.1M tweets for 2.1M users.

%%%%%%%%%%%%%%%%%%
\subsection{Annotation}\label{sec:data:annotation} 
%%%%%%%%%%%%%%%%%%
After collection, we annotated user characteristics. More specifically, we created reference labels for gender, age, and location.
Here, we briefly outline the annotation of these characteristics. %More details can be found in \Cref{sec:appendix_data}. 
First, we coded the (binary) gender of Italian Twitter users based on the full name field of each Twitter user. This process was facilitated by the fact that Italian names are gender-specific in nearly all cases.\footnote{We used the most common Italian names and their associated gender reported on this list: \url{https://github.com/mrblasco/genderNamesITA}}
Since many users share their full names on Twitter, we were able to assign reference labels to 1.3M users out of 2.1M.
Second, we annotated the age of users, constructing a set of regex patterns matching users' statements of birthyear or current age in either their bio or any of their tweets.
Differently from the gender annotation, we were able to assign reference labels for age only to a minor fraction of the users: 33K users.
This imbalance highlights an important aspect of the demographic classification task on Twitter: there is a strong need for age classifiers for the vast majority of users, while gender can be classified for large shares of the social media population with rule-based classifiers, or even with a simple dictionary, as we did.

Lastly, we identified each profile's approximate locations by matching the profile location information against a list of Italian regions and municipalities.
All three procedures for annotating reference labels were designed to achieve high accuracy at the expense of potentially annotating fewer users. Nonetheless, we are left with close to 20K users for which we have reference labels for both gender and age. In the final dataset, we focus on users with both gender and age information to have a consistent sample for the prediction across models. 

%In the final dataset, we prioritize users with both gender and age information since these are features that we can try predicting with text-based models when explicit information (the one used to code reference labels) is not available.\footnote{Although some literature on location prediction based on users' bio and tweets exists, the task seem very hard \citep{}}

\subsubsection{Data quality}

To ensure the accuracy of the coding, three members of the research team manually verified the reference labels for gender and age in a random subset of 1,200 users for gender and 350 users for age. For gender, the three annotators marked between 99.33\% and 99.42\% of the labels as correct, with an average inter-annotator raw agreement of 99.94\%. For age, the annotators marked between 93.55\% and 97.10\% of the labels as correct, with an average inter-annotator raw agreement of 94.41\%.  Two out of three annotators extended the verification of age labels to all the 1,200 users that were verified for gender and then discarded from the dataset the users with incorrect labels (either gender or age) according to at least one human annotator, keeping the remaining as a test set for the following experiments on gender and age classification. Hence, our test set guarantees 100\% quality (according to two annotators) labels for each user.

To make the data even more useful for follow-up research, we also evaluate the representativeness of our data relative to the population of Italian Twitter users. Overall, the results, which we discuss and visualize in greater detail in \Cref{sec:appendix_data}, suggest that our data closely mimic the characteristics of the overall Italian Twitter population.

% Gender manual eval:
% Good: 1197 (99.75\%)
% Bad: 3 (0.25\%)

% Age manual eval:
% Good: 45 (96.20\%)
% Bad: 1139 (3.80\%)

%%%%%%%%%%%%%%%%%%%%%%%%%%%%%%%%%%%%%%%%%%%%%%%%%%%%%%%%%%%%%
\begin{table}[htb!]
    \centering
    \footnotesize
    \begin{tabular}{lccc}
    \toprule
    & \textbf{Train} & \textbf{Valid} & \textbf{Test} \\
    \midrule
    \# Users & 17262& 1919& 1119\\
    \hspace{0.2cm}\% w/ Pic & 0.80& 0.78& 1.00\\
    \hspace{0.2cm}\% w/ Bio & 0.47& 0.45& 0.46\\
    \hspace{0.2cm}\% w/ Location & 43.0& 41.1& 47.2\\
    \hspace{0.2cm}\% Female & 0.37& 0.37& 0.36\\
    Average Age & 42.3& 42.7& 42.3\\
    \# Tweets & 24.5M& 2.8M& 1.6M\\
    \hspace{0.2cm}\% RT & 23.8& 24.6& 19.5\\
    \# Tweets/User & 338& 355& 274\\
    \bottomrule
    \end{tabular} 
    \caption{DADIT overview \label{tab:data}}
\end{table}
%%%%%%%%%%%%%%%%%%%%%%%%%%%%%%%%%%%%%%%%%%%%%%%%%%%%%%%%%%%%%

%%%%%%%%%%%%%%%%%%%%%%%%%%%%%%%%%%%%%%%%%%%%%%%%%%%%%%%%%%%%%%%%%%%%%%%%%%%%%%%%%%%%%%%%%%%%%%
%%%%%%%%%%%%%%%%%%%%%%%%%%%%%%%%%%%%%%%%%%%%%%%%%%%%%%%%%%%%%%%%%%%%%%%%%%%%%%%%%%%%%%%%%%%%%%
\section{Gender and Age Classification\label{sec:exp}}
%%%%%%%%%%%%%%%%%%%%%%%%%%%%%%%%%%%%%%%%%%%%%%%%%%%%%%%%%%%%%%%%%%%%%%%%%%%%%%%%%%%%%%%%%%%%%%
%%%%%%%%%%%%%%%%%%%%%%%%%%%%%%%%%%%%%%%%%%%%%%%%%%%%%%%%%%%%%%%%%%%%%%%%%%%%%%%%%%%%%%%%%%%%%%

In this section, we analyze whether  1) tweets constitute a valuable source of information for the classification of the gender and age of users; 2) text-only classifiers can compete with standard multimodal models like M3; 3) all the modalities available in our dataset (text and pictures) can be leveraged in a simple way to improve performance. 

%%%%%%%%%%%%%%%%%%
\subsection{Experimental setting}\label{sec:exp:setting}
%%%%%%%%%%%%%%%%%%

We use the constructed dataset to train and compare a number of gender and age classifiers.
For our analysis, we divide our database of 20K Twitter users into a training (85.0\%), validation (9.5\%), and test set (5.5\%) set. 
\Cref{tab:data} provides an overview of each split's users, tweets, and demographic characteristics. Note that we split the data to ensure that all users in the test set have profile images. Otherwise, all samples are identical based on demographic characteristics.

For the classification task, we are interested in cases where gender and age can not be inferred with the rule-based methods that we used for the automatic annotation of DADIT, because of the lack of the necessary clues.
We simulate this situation by avoiding to provide models with users' full names as a feature, and we mask age and birthyear information from bios and tweets.
We apply a light preprocessing to tweets: removing mentions, URLs, and any tweet shorter than three words. 
For each user, we concatenate their bio and most recent tweets (until the maximum input length is reached) to form one contiguous text to be input as features in the text classifiers. We compare the performance of a dummy classifier that predicts the majority class with four models.

%%%%%%%%%%%%%%%%%%
\textbf{M3}\hspace{12pt} 
%%%%%%%%%%%%%%%%%%
As a strong baseline, we use the approach of \newcite{wang2019demographic}, which is one of the standard models for inferring the age and gender characteristics of Twitter users \cite{bianchi2022twitter}.
M3 can use users' usernames, biographies, and profile images as features, but not tweets. In our analysis, we avoid providing usernames as we experiment in a setting where this information is unavailable (otherwise, gender could be inferred with high confidence as we did in Section \ref{sec:data:annotation}).
M3 classifies ages into four classes: (0-19), [19-30), [30-40), and [40-100). We adopt the same age classes for all the other models to make them comparable.
We compare the performance of M3 with and without the profile image. Omitting the profile image tells us about M3's performance when users do not provide a profile image or the image offers no identifiable characteristics.

%%%%%%%%%%%%%%%%%%
\textbf{CV}\hspace{12pt} 
%%%%%%%%%%%%%%%%%%
As an alternative baseline classifier, we use another popular, open-source model based on computer vision (CV) \cite{levi2015age}. This model predicts gender and age solely based on users' profile pictures.

%%%%%%%%%%%%%%%%%%
\textbf{XLM}\hspace{12pt} 
%%%%%%%%%%%%%%%%%%
We leverage our dataset to train a classifier based on \texttt{twitter-XLM-roberta-base} \cite{barbieri2021xlm}\footnote{\url{https://huggingface.co/cardiffnlp/twitter-xlm-roberta-base}}, a multilingual language model finetuned on twitter data. We train a classification head for gender and age, separately, and we also finetune the weights of the backbone model.
%Mahyar please add info in appendix about n. layers, hyperparameters,epochs,stopping criterion, max_input_len...

%%%%%%%%%%%%%%%%%%
\textbf{Flan-T5}\hspace{12pt} 
%%%%%%%%%%%%%%%%%%
Flan-T5 is a multilingual, instruction-finetuned language model achieving state-of-the-art performance in several zero-shot tasks \cite{chung2022scaling}. We test its largest checkpoint (\texttt{flan-t5-xxl})\footnote{\url{https://huggingface.co/google/flan-t5-xxl}} in a zero-shot classification setting, as in \citet{plaza2023leveraging}.
We use it to classify the gender and age of users based on simple prompts reported in Appendix \ref{sec:appendix_exp}.

%%%%%%%%%%%%%%%%%%
\textbf{GPT3.5}\hspace{12pt} 
%%%%%%%%%%%%%%%%%%
Finally, we test the \texttt{gpt-3.5-turbo} through OpenAI's API, in a few-shot setting. GPT3.5 is one of the largest LMs available to date, the successor of the popular GPT3 \cite{brown2020language}.
We use it to classify the gender and age of users based on simple prompts reported in Appendix \Cref{sec:appendix_exp}. In our preliminary experiments, GPT performed consistently better when provided with labeled examples in the prompt, consistently with \citet{lupo2023use}. Therefore, we test it in a few-shot setting with two examples for each age class, and five for each gender.

% \usepackage[normalem]{ulem}
\begin{table*}[htb!]
\footnotesize
\centering
\begin{tabular}{llcc|cc||cc|cc}
 \toprule
 &  & \multicolumn{4}{c}{\textbf{Italian}} & \multicolumn{4}{c}{\textbf{German}} \\
 &  & \multicolumn{2}{c}{\textbf{Gender}} & \multicolumn{2}{c}{\textbf{Age}} & \multicolumn{2}{c}{\textbf{Gender}} & \multicolumn{2}{c}{\textbf{Age}} \\
Model & Features & F1 & Acc & F1 & Acc & F1 & Acc & F1 & Acc \\
 \midrule
Dummy & majority class & 39.2 & 64.4 & 15.9 & 46.8 & 44.1 & 78.9 & 17.0 & 51.3 \\
M3 & bio & 50.9 & 67.7 & 16.6 & 16.4 & - & - & - & - \\
M3 & bio+pic & 79.3 & 83.1 & 41.4 & 53.7 & 56.2 & 80.2 & 25.9 & 31.7 \\
CV & pic & 70.7 & 76.0 & 27.7 & 35.2 & 64.7 & 79.4 & 28.2 & 35.6 \\
XLM & bio & 81.6 & 72.5 & 33.3 & 52.2 & - & - & - & - \\
XLM & bio+twt & \ul{86.8} & \ul{88.1} & \ul{\textbf{60.2}} & \ul{66.4} & \ul{74.3} & 82.0 & \ul{45.8} & \ul{58.0} \\
Flan-T5 & bio & 62.1 & 70.5 & 16.3 & 16.2 & - & - & - & - \\
Flan-T5 & bio+twt & 75.4 & 76.8 & 34.0 & 58.0 & \ul{\textbf{80.8}} & \ul{\textbf{87.4}} & 29.0 & 35.1 \\
GPT3.5 & bio & 64.9 & 72.8 & 21.1 & 20.3 & - & - & - & - \\
GPT3.5 & bio+twt & 77.4 & 78.0 & 31.4 & 32.0 & 48.2 & 72.2 & 31.4 & \ul{41.2} \\
XLM+M3 & bio+twt+pic-
& \ul{\textbf{90.8}} & \ul{\textbf{91.9}} & \ul{60.0} & \ul{\textbf{67.2}} & \ul{74.8} & 83.6 & \ul{\textbf{48.1}} & \ul{\textbf{58.8}} \\
\bottomrule
\end{tabular}
    \caption{Comparison of gender and age classifiers on Italian and German Twitter users. Highest performance per column in bold. Results outperforming our strong baseline (M3 bio+pic) with statistical significance (p<0.01) are underlined.}\label{tab:model_performance}
\end{table*}

\subsection{Results \label{sec:results}}
%%%%%%%%%%%%%%%%%%%%%%%%%%%%%%%%%%%%%%%%%%%%%%%%%%%%%%%%%%%%%%%%%%%%%%%%%%%%%%%%%%%%%%%%%%%%%%
%%%%%%%%%%%%%%%%%%%%%%%%%%%%%%%%%%%%%%%%%%%%%%%%%%%%%%%%%%%%%%%%%%%%%%%%%%%%%%%%%%%%%%%%%%%%%%

The results of the models' evaluation on our Italian test set are in Tables \ref{tab:model_performance}. Three main findings stand out. First, we document that including information from tweets consistently and significantly improves model performance. For example, the F1-scores for gender prediction of both Flan-T5 and GPT3.5 increase by over 10 points. We observe even larger performance gains for the prediction of age.

Second, we find that the finetuned XLM stands out as the top performer in both tasks, particularly for age prediction, which is most crucial for users' demographic classification, as discussed earlier. Even with profile images, M3 achieves an F1-score of 41.4 relative to an F1-score of 60.2 of the finetuned model. The F1-score of the finetuned model is also nearly twice as high as the score achieved by other competitors. 

Third, we show that additional performance gain for the prediction of gender can be achieved by a simple ensemble learner that combines the predictions from M3 and XLM, weighting their output probabilities with their F1-scores. Surprisingly, however, ensemble results do not improve over XLM on age prediction. 

Lastly, we compare the best Italian classifiers on a German test set, collected and annotated following the same procedure described in \Cref{sec:data:annotation}.\footnote{More details on the German test set in Appendix \ref{sec:appendix_data}.} Note that, after fine-tuning XLM on the Italian training data, we do not further finetune it on any German training data, since we lack them. Instead, we leverage XLM's multilingual abilities to solve the German classification tasks without further fine-tuning. 
The findings from the German test set validate our results in Italian.
Interestingly, we find that even though XLM was only finetuned on Italian Twitter data, it is still among the best-performing models for gender prediction in German, and achieves the highest F1-score in the prediction of age.
This finding highlights again the importance of annotated datasets like DADIT.

%%%%%%%%%%%%%%%%%%%%%%%%%%%%%%%%%%%%%%%%%%%%%%%%%%%%%%%%%%%%%
% \begin{table}[htb!]
%     \centering
%     \footnotesize
%     \begin{tabular}{llccccc}
%     \toprule
%  &  & \multicolumn{2}{c}{Gender} & \multicolumn{2}{c}{Age} \\
% Model & Features & F1 & Acc & F1 & Acc \\
%  \midrule
% Dummy & majority & 44.1 & 78.9 & 17.0 & \underline{51.3} \\
% M3 & bio+pic & 56.2 & 80.2 & 25.9 & 31.7 \\
% CV & pic & 64.7 & 79.4 & 28.2 & 35.6 \\
% XLM & bio+twt & \underline{74.3} & 82.0 & \underline{\textbf{45.8}} & \underline{\textbf{58.0}} \\
% GPT3.5 & bio+twt & 48.2 & 72.2 & 31.4 & \underline{41.2} \\
% FLAN-T5 & bio+twt & \underline{\textbf{80.8}} & \underline{\textbf{87.4}} & 29.0 & 35.1 \\
%     \bottomrule
%     \end{tabular}
%     \caption{Model Performance Comparison on German Data}\label{tab:model_performance_de}
% \end{table}
%%%%%%%%%%%%%%%%%%%%%%%%%%%%%%%%%%%%%%%%%%%%%%%%%%%%%%%%%%%%%
%%%%%%%%%%%%%%%%
\subsection{Scaling up}

In this section we explore ways to push even further the performance of our age and gender classification approach, with a focus on the Italian dataset.
Two strategies are evaluated: employing a larger pre-trained encoder for the classifier and incorporating additional user-level features.
We transition from \texttt{twitter-XLM-roberta-base} to the larger \texttt{twitter-XLM-roberta-large-2022} (XLM-large - \citet{barbieri2021xlm}\footnote{\url{https://huggingface.co/cardiffnlp/twitter-xlm-roberta-large-2022}} ).
Table \ref{tab:model_scaled} outlines the observed changes in classification performance, revealing a modest enhancement in age prediction and a more notable improvement in gender classification.

We next incorporate extra user-level features as input to our classifiers. In the age classifier, we added users' names, their Twitter usernames, account creation years, and counts of tweets, followings, and followers. For gender classification, we used all these features but omitted users' names for the reasons explained above.\footnote{The gender classifier trained with users' names achieves nearly perfect accuracy, as expected} The inclusion of these features markedly improved the accuracy of age classification. However, the gender classification model showed a minor decline in performance.

% \begin{table}[]
% \begin{tabular}{lccccc}
% \toprule
% Model & Target & Features & Groups & F1 & Acc \\
% \midrule
% XLM & Age &bio+twt & 4 & 60.2 & 66.4 \\
% XLM-large & Age & bio+twt & 4 & 61.1 & 67.3 \\
% XLM-large & Age & all & 4 & \textbf{63.5} & \textbf{69.9} \\
% XLM-large & Age & all & 5 & 61.3 & 61.1 \\
% \bottomrule
% \end{tabular}
% \caption{Age prediction with Italian XLM classifiers}
% \end{table}

\begin{table}[]
\begin{tabular}{lcccc}
\toprule
Model & Target & Features & F1 & Acc \\
\midrule
XLM & Age & bio+twt  & 60.2 & 66.4 \\
XLM-large & Age & bio+twt  & 61.1 & 67.3 \\
XLM-large & Age & all  & \textbf{63.5} & \textbf{69.9} \\
\midrule
XLM & Gender & bio+twt & 86.8 & 88.1 \\
XLM-large & Gender & bio+twt & \textbf{93.0} & \textbf{90.9} \\
XLM-large & Gender & all-name & 91.8 & 89.0 \\
%XLM & Gender
\bottomrule
\end{tabular}
\caption{Age and gender prediction with XLM classifiers on the Italian dataset.} \label{tab:model_scaled}
\end{table}

%%%%%%%%%%%%%%%%%%%%%%%%%%%%%%%%%%%%%%%%%%%%%%%%%%%%%%%%%%%%%
%%%%%%%%%%%%%%%%
\subsection{Quality-Coverage Tradeoff}
%%%%%%%%%%%%%%%%%%%%%%%%%%%%%%%%%%%%%%%%%%%%%%%%%%%%%%%%%%%%%

%%%%%%%%%%%%%%%%%%%%%%%%%%%%%%%%%%%%%%%%%%%%%%%%%%%%%%%%%%%
\begin{figure*}[htb!]
    \centering
    \includegraphics[width=0.8\textwidth]{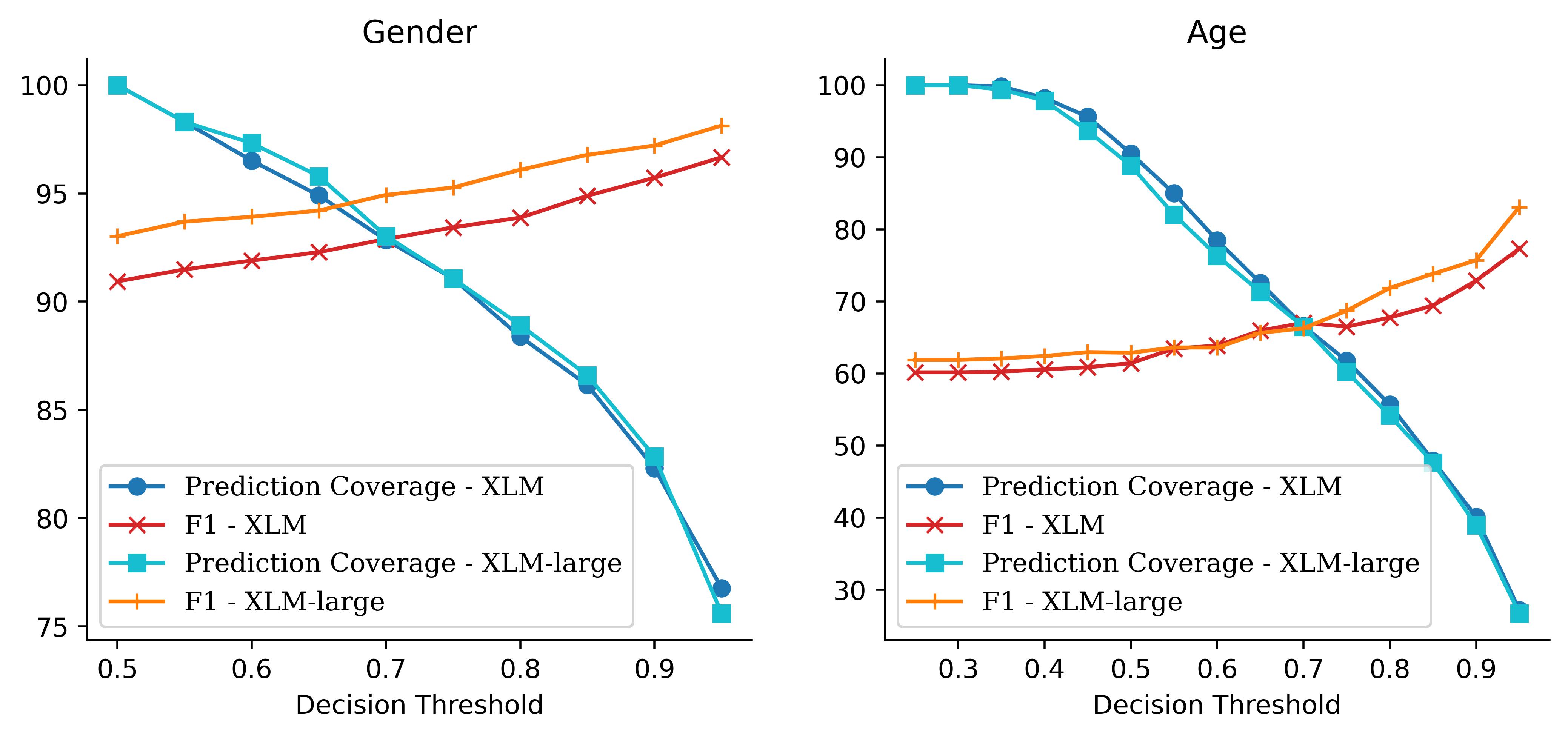} 
    \caption{F1 score and prediction coverage tradeoff. \label{fig:tradeoff}}  
\end{figure*}
%%%%%%%%%%%%%%%%%%%%%%%%%%%%%%%%%%%%%%%%%%%%%%%%%%%%%%%%%%%%%

For settings requiring higher classification quality, we can comply at the expense of some coverage of classified users by restricting the classification to those users that the model classifies with higher confidence.
Typically, observations are classified based on the highest probability class output by the model.
Therefore, we can restrict the classification to the subset of users where the model's output probability for the most likely class exceeds a decision threshold.
In Figure \ref{fig:tradeoff}, we show that the decision threshold (minimum probability for the predicted class) is positively correlated to the F1 score but negatively correlated to the share of users that can be predicted, for both gender and age classification.
For instance, in gender classification, setting a threshold of 0.8 improves the F1 score by approximately 3 points against the standard 0.5 threshold.\footnote{We used this decision threshold, which is equivalent to using none, in all our previous experiments on gender.}
However, the 0.8 threshold also reduces the classification coverage of users by over 10\%. In other words, more than 10\% of the users can not be classified at the given confidence level.
In age classification, the same threshold boosts the F1 score by around 5 percentage points, yet it halves the coverage of classified users.

%%%%%%%%%%%%%%%%
\section{Related Work}
Accurately predicting the demographic characteristics of social media users has become central for computational social science. The wealth of information on online behavior has been used to infer, among others, demographic characteristics, income, political ideology, religion, and health outcomes \cite{al2012homophily,culotta2014estimating,barbera2015birds,preoctiuc2017beyond,preoctiuc2018user,matz2019predicting,jaidka2020estimating,fujiwara2023effect,chaturvedi2023s}. Especially the demographic characteristics of users (e.g., gender, age, race, ethnicity) are used to characterize heterogeneity in the behavior of users groups (e.g., women and men) \cite[e.g.,][]{holmberg2015gender,usher2018twitter,bathina2021individuals,zhang2021monitoring,burel2021demographics,allen2022birds}. Further, it has been shown that demographic characteristics can improve classification performance on other attributes \cite{volkova2013exploring,hovy2015demographic,lynn2017human,li2018towards,hovy2021importance}. As a result, models of user classification have become a stable of modern social media analysis. % \cite{vempala2019categorizing}

Existing approaches to classify social media user data encompass various techniques that leverage the information available in online interactions. Overall, three main approaches are used for the classification of users. First and most frequently, information is extracted from the user profile information (e.g., screen name, bio) \cite{wang2019demographic,lockhart2023name}. Secondly, the profile information can be enriched using information from the profile image \cite{alowibdi2013language,wang2019demographic}. The profile image is particularly informative for classifying users' gender, age, and ethnicity. A third approach is analyzing the network connections on the social media platform. These approaches have been particularly successful in classifying the age and political ideology of users\cite{al2012homophily,barbera2015birds}.

%We provide a novel Twitter data set to study inequality enriched with high-quality demographic data. 

%%%%%%%%%%%%%%%%%%%%%%%%%%%%%%%%%%%%%%%%%%%%%%%%%%%%%%%%%%%%%%%%%%%%%%%%%%%%%%%%%%%%%%%%%%%%%%
%%%%%%%%%%%%%%%%%%%%%%%%%%%%%%%%%%%%%%%%%%%%%%%%%%%%%%%%%%%%%%%%%%%%%%%%%%%%%%%%%%%%%%%%%%%%%%
\section{Conclusion \label{sec:conclusion}}
%%%%%%%%%%%%%%%%%%%%%%%%%%%%%%%%%%%%%%%%%%%%%%%%%%%%%%%%%%%%%%%%%%%%%%%%%%%%%%%%%%%%%%%%%%%%%%
%%%%%%%%%%%%%%%%%%%%%%%%%%%%%%%%%%%%%%%%%%%%%%%%%%%%%%%%%%%%%%%%%%%%%%%%%%%%%%%%%%%%%%%%%%%%%%

Social media platforms have provided a goldmine of data for machine learning and NLP models to analyze user behaviour and interactions. We introduce DADIT: a rigorously curated, demographically annotated data set for the social sciences. It contains over 20K representative Italian Twitter users, with over 30M tweets, useful for the study of political attitudes and beliefs. As social media analysis usually relies on the demographic characteristics of the users for stratification, we compare the performance of different methods for the automated prediction of users' gender and age in contexts where this information can not be easily retrieved from users' profiles.
We show the latent potential of tweet content to enrich user classification approaches for all tweeters, regardless of profile information. Second, we demonstrate that finetuned models based on user bios and tweets systematically outperform multimodal state-of-the-art methods based on bios and profile images. Lastly, we suggest a lightweight solution to take advantage of the multimodality of our data.

%Numerous research have shown that automated work. However, we analyze the reliability and find irregularities. As biographical information and profile pictures are missing or absent, existing models struggle. We focus on tweet text for user classification, improving accuracy and predictability.

%Our study shows our strategy works. Through careful comparisons with state-of-the-art models using profile information, we show that our tweet-based model meets or exceeds many benchmarks. Additionally, our solution consistently improves previous approaches across a variety of parameters. We conclude that our approach advances user categorisation. It advances the field toward a more inclusive, data-rich future where textual material and user profiles enhance our knowledge of social media interactions.

%%%%%%%%%%%%%%%%%%%%%%%%%%%%%%%%%%%%%%%%%%%%%%%%%%%%%%%%%%%%%%%%%%%%%%%%%%%%%%%%%%%%%%%%%%%%%%
%%%%%%%%%%%%%%%%%%%%%%%%%%%%%%%%%%%%%%%%%%%%%%%%%%%%%%%%%%%%%%%%%%%%%%%%%%%%%%%%%%%%%%%%%%%%%%
\section*{Acknowledgements }
%%%%%%%%%%%%%%%%%%%%%%%%%%%%%%%%%%%%%%%%%%%%%%%%%%%%%%%%%%%%%%%%%%%%%%%%%%%%%%%%%%%%%%%%%%%%%%
%%%%%%%%%%%%%%%%%%%%%%%%%%%%%%%%%%%%%%%%%%%%%%%%%%%%%%%%%%%%%%%%%%%%%%%%%%%%%%%%%%%%%%%%%%%%%%

The authors are grateful for the financial support of the Cariplo Foundation as part of the MENTALISM Project (MEasuriNg, Tracking, and AnaLyzing Inequality using Social Media). We are grateful to Derya Durmush for her excellent research assistance. We also thank Debora Nozza for her insightful comments. We acknowledge the CINECA award under the ISCRA initiative, for the availability of high-performance computing resources and support. Dirk Hovy is a member of the Bocconi Institute for Data Science and Analytics (BIDSA). Carlo Schwarz is a member of the PERICLES and IGIER research centres at Bocconi as well as a Research Affiliate of the CEPR.

%%%%%%%%%%%%%%%%%%%%%%%%%%%%%%%%%%%%%%%%%%%%%%%%%%%%%%%%%%%%%%%%%%%%%%%%%%%%%%%%%%%%%%%%%%%%%%
%%%%%%%%%%%%%%%%%%%%%%%%%%%%%%%%%%%%%%%%%%%%%%%%%%%%%%%%%%%%%%%%%%%%%%%%%%%%%%%%%%%%%%%%%%%%%%
\section*{Ethical Considerations}
\label{sec:ethics}
%%%%%%%%%%%%%%%%%%%%%%%%%%%%%%%%%%%%%%%%%%%%%%%%%%%%%%%%%%%%%%%%%%%%%%%%%%%%%%%%%%%%%%%%%%%%%%
%%%%%%%%%%%%%%%%%%%%%%%%%%%%%%%%%%%%%%%%%%%%%%%%%%%%%%%%%%%%%%%%%%%%%%%%%%%%%%%%%%%%%%%%%%%%%%

We anonymized Twitter handles as part of the preprocessing to ensure GDPR compliance with at least pseudo-anonymity (motivated intruders are still likely to be able to recover identities with some effort, given that the data is in the public domain).
In the reproduction data, tweet IDs are only shared in the dehydrated form in compliance with Twitter's Developer Policies: ``Individuals redistributing tweet IDs and/or User IDs on behalf of an academic institution for the sole purpose of non-commercial research are permitted to redistribute an unlimited number of tweet IDs and/or User IDs.''
\url{https://developer.twitter.com/en/developer-terms/more-on-restricted-use-cases}. 
Hydrated and non-anonymized data will only be made available for research purposes upon request. We are aware that the non-anonymized data could be used to target individuals. As all data are already in the public domain, the additional risk is minimal, and we believe that the value of the data for the scientific community outweighs the risk.

%%%%%%%%%%%%%%%%%%%%%%%%%%%%%%%%%%%%%%%%%%%%%%%%%%%%%%%%%%%%%%%%%%%%%%%%%%%%%%%%%%%%%%%%%%%%%%
%%%%%%%%%%%%%%%%%%%%%%%%%%%%%%%%%%%%%%%%%%%%%%%%%%%%%%%%%%%%%%%%%%%%%%%%%%%%%%%%%%%%%%%%%%%%%%
\section*{Limitations }
%%%%%%%%%%%%%%%%%%%%%%%%%%%%%%%%%%%%%%%%%%%%%%%%%%%%%%%%%%%%%%%%%%%%%%%%%%%%%%%%%%%%%%%%%%%%%%
%%%%%%%%%%%%%%%%%%%%%%%%%%%%%%%%%%%%%%%%%%%%%%%%%%%%%%%%%%%%%%%%%%%%%%%%%%%%%%%%%%%%%%%%%%%%%%

There are two minor limitations to the external validity of our findings. First, while we confirm the representativeness of our dataset relative to the overall Italian Twitter population, our sample is likely not a random set of Italian Twitter users. Further, as we label gender based on names, our gender coding is binary and might not fully represent the gender expression of users. We nonetheless believe that the DADIT data hold great promise for the study of attitudes and beliefs of the Italian population, as Twitter data are never representative of the overall population in any case. 

A second limitation regards the multimodal potential of DADIT: we trained text classifiers on it, but only ran pre-trained vision classifiers in inference mode. Finetuning vision models on our data could bring further classification improvements, although likely only to a limited extent, as they have already been trained on larger datasets. Perhaps more interestingly, future works could achieve performance improvements with a stronger ensemble method, i.e., training a classification head on top of the image and text representations output by the vision and text models.

\section*{Data and Code availability}
%%%%%%%%%%%%%%%%%%%%%%%%%%%%%%%%%%%%%%%%%%%%%%%%%%%%%%%%%%%%%%%%%%%%%%%%%%%%%%%%%%%%%%%%%%%%%%
%%%%%%%%%%%%%%%%%%%%%%%%%%%%%%%%%%%%%%%%%%%%%%%%%%%%%%%%%%%%%%%%%%%%%%%%%%%%%%%%%%%%%%%%%%%%%%

We publicly release DADIT and our best gender and age classifiers though Hugging Face, at this address: \url{https://huggingface.co/lorelupo}. The software to train and run the classifiers, or to replicate our results is available at \url{https://github.com/lorelupo/twitter_user_classification}.

%%%%%%%%%%%%%%%%%%%%%%%%%%%%%%%%%%%%%%%%%%%%%%%%%%%%%%%%%%%%%%%%%%%%%%%%%%%%%%%%%%%%%%%%%%%%%%
%%%%%%%%%%%%%%%%%%%%%%%%%%%%%%%%%%%%%%%%%%%%%%%%%%%%%%%%%%%%%%%%%%%%%%%%%%%%%%%%%%%%%%%%%%%%%%
\section{Bibliographical References}\label{sec:reference}
\bibliographystyle{lrec-coling2024-natbib}
\bibliography{bib_MENTALISM}
%%%%%%%%%%%%%%%%%%%%%%%%%%%%%%%%%%%%%%%%%%%%%%%%%%%%%%%%%%%%%%%%%%%%%%%%%%%%%%%%%%%%%%%%%%%%%%
%%%%%%%%%%%%%%%%%%%%%%%%%%%%%%%%%%%%%%%%%%%%%%%%%%%%%%%%%%%%%%%%%%%%%%%%%%%%%%%%%%%%%%%%%%%%%%

%%%%%%%%%%%%%%%%%%%%%%%%%%%%%%%%%%%%%%%%%%%%%%%%%%%%%%%%%%%%%%%%%%%%%%%%%%%%%%%%%%%%%%%%%%%%%%
%%%%%%%%%%%%%%%%%%%%%%%%%%%%%%%%%%%%%%%%%%%%%%%%%%%%%%%%%%%%%%%%%%%%%%%%%%%%%%%%%%%%%%%%%%%%%%
%%%%%%%%%%%%%%%%%%%%%%%%%%%%%%%%%%%%%%%%%%%%%%%%%%%%%%%%%%%%%%%%%%%%%%%%%%%%%%%%%%%%%%%%%%%%%%
%%%%%%%%%%%%%%%%%%%%%%%%%%%%%%%%%%%%%%%%%%%%%%%%%%%%%%%%%%%%%%%%%%%%%%%%%%%%%%%%%%%%%%%%%%%%%%
\newpage
\clearpage
\appendix
%%%%%%%%%%%%%%%%%%%%%%%%%%%%%%%%%%%%%%%%%%%%%%%%%%%%%%%%%%%%%%%%%%%%%%%%%%%%%%%%%%%%%%%%%%%%%%
%%%%%%%%%%%%%%%%%%%%%%%%%%%%%%%%%%%%%%%%%%%%%%%%%%%%%%%%%%%%%%%%%%%%%%%%%%%%%%%%%%%%%%%%%%%%%%
%%%%%%%%%%%%%%%%%%%%%%%%%%%%%%%%%%%%%%%%%%%%%%%%%%%%%%%%%%%%%%%%%%%%%%%%%%%%%%%%%%%%%%%%%%%%%%
%%%%%%%%%%%%%%%%%%%%%%%%%%%%%%%%%%%%%%%%%%%%%%%%%%%%%%%%%%%%%%%%%%%%%%%%%%%%%%%%%%%%%%%%%%%%%%

%%%%%%%%%%%%%%%%%%%%%%%%%%%%%%%%%%%%%%%%%%%%%%%%%%%%%%%%%%%%%%%%%%%%%%%%%%%%%%%%%%%%%%%%%%%%%%
%%%%%%%%%%%%%%%%%%%%%%%%%%%%%%%%%%%%%%%%%%%%%%%%%%%%%%%%%%%%%%%%%%%%%%%%%%%%%%%%%%%%%%%%%%%%%%
\section{Dataset Details \label{sec:appendix_data}}
%%%%%%%%%%%%%%%%%%%%%%%%%%%%%%%%%%%%%%%%%%%%%%%%%%%%%%%%%%%%%%%%%%%%%%%%%%%%%%%%%%%%%%%%%%%%%%
%%%%%%%%%%%%%%%%%%%%%%%%%%%%%%%%%%%%%%%%%%%%%%%%%%%%%%%%%%%%%%%%%%%%%%%%%%%%%%%%%%%%%%%%%%%%%%

\paragraph{Italian Data}
The data collection of the Italian Twitter data started from a list of all Italian senators and deputies in the XVIII Legislature (2018-2022). We manually searched for Twitter accounts for each of the politicians. It is worth noting that a few politicians do not have a Twitter account, while less than a handful have more than one account associated with their name. In total, we obtained Twitter accounts for 802 Italian politicians. \Cref{tab:source_list} provides an overview of the number of accounts that were collected for each of the Italian parties. In total, these accounts are followed by 3.2 million unique users, for which we then collected user-profiles and tweets.

\begin{table}[htbp]
    \centering
      \footnotesize
    \begin{tabular}{lcc}
    \toprule
    \textbf{Party} & \textbf{\# Accounts} & \textbf{\# Followers} \\
    \midrule
Movimento 5 Stelle                           & 191   & 2,548,834\\
Lega                                         & 172   & 2,413,691\\
Partito Democratico                          & 131   & 5,209,329\\
Forza Italia                                 &  97   & 1,441,865\\
Misto                                        &  74   & 3,013,421\\
Fratelli d'Italia                            &  55   & 1,526,940\\
Italia Viva                                  &  39   & 4,728,160\\
Coraggio Italia                              &  16   &   29,787\\
Liberi e Uguali                              &   8   & 1,051,836\\
Indipendente                                 &   7   &  479,298\\
C.A.L.                                       &   6   &   32,622\\
% (costituzione, ambiente, lavoro)
Per le Autonomie                             &   3   &  166,589\\
Azione                                       &   1   &  349,375\\
Europa Verde                                 &   1   &    9,969\\
Südtiroler Volkspartei                       &   1   &    3,033\\   
\bottomrule
\end{tabular} 
\caption{Summary of Italian source accounts. \label{tab:source_list}}
\end{table}

\paragraph{German Data}
The German dataset results from a similar data collection as the Italian data. The authors collected the Twitter followers of eight German parties (CDU, CSU, SPD, FDP, Greens, The Left, AfD, NPD). \Cref{tab:source_list_de} lists the number of followers for each of the collected accounts. We then collected the user profiles and tweets of the followers. In total, we obtained around 965,000 user profiles. Afterward, we label users' gender based on the "full name" and age based on the user bios. This leaves us with 388 user profiles for which we have both gender and age coding. For this subsample, we then collected user images.  

\begin{table}[ht]
    \centering
      \footnotesize
    \begin{tabular}{lc}
    \toprule
    \textbf{Party} &\textbf{\# Followers} \\
    \midrule
CDU                      & 297,417\\
CSU                      & 196,119\\
SPD                      & 370,956\\
FDP                      & 350,234\\
Greens                   & 488,959\\
The Left                 & 276,713\\
AfD                      & 147,942\\
NPD                      & 5,184\\
\bottomrule
\end{tabular} 
\caption{Summary of German source accounts. \label{tab:source_list_de}}
\end{table}

%%%%%%%%%%%%%%%%%%
\subsection{Representativness \label{sec:representativnes}}
%%%%%%%%%%%%%%%%%%

We evaluate the representativeness of the Italian data relative to the demographic characteristics of Italian Twitter users. Unfortunately, there is no census of Italian Twitter users from which we could obtain the gender and age of the universe of Italian Twitter users. Instead, we rely on information obtained from research by \citet{alessandra2019tweets} and a report from \citet{statista_twitter}. The former provides information on the gender and location of Twitter users, while the latter contains information on the age of the Twitter user base. As the report by \citet{statista_twitter} splits the age of users by generation, we convert the age of the users to the same generations. Note that these generations differ from the age bins we use in the main paper.  

\begin{figure*}[ht]
\centering
    \begin{subfigure}{\linewidth}
        \centering
        \includegraphics[width=0.9\linewidth]{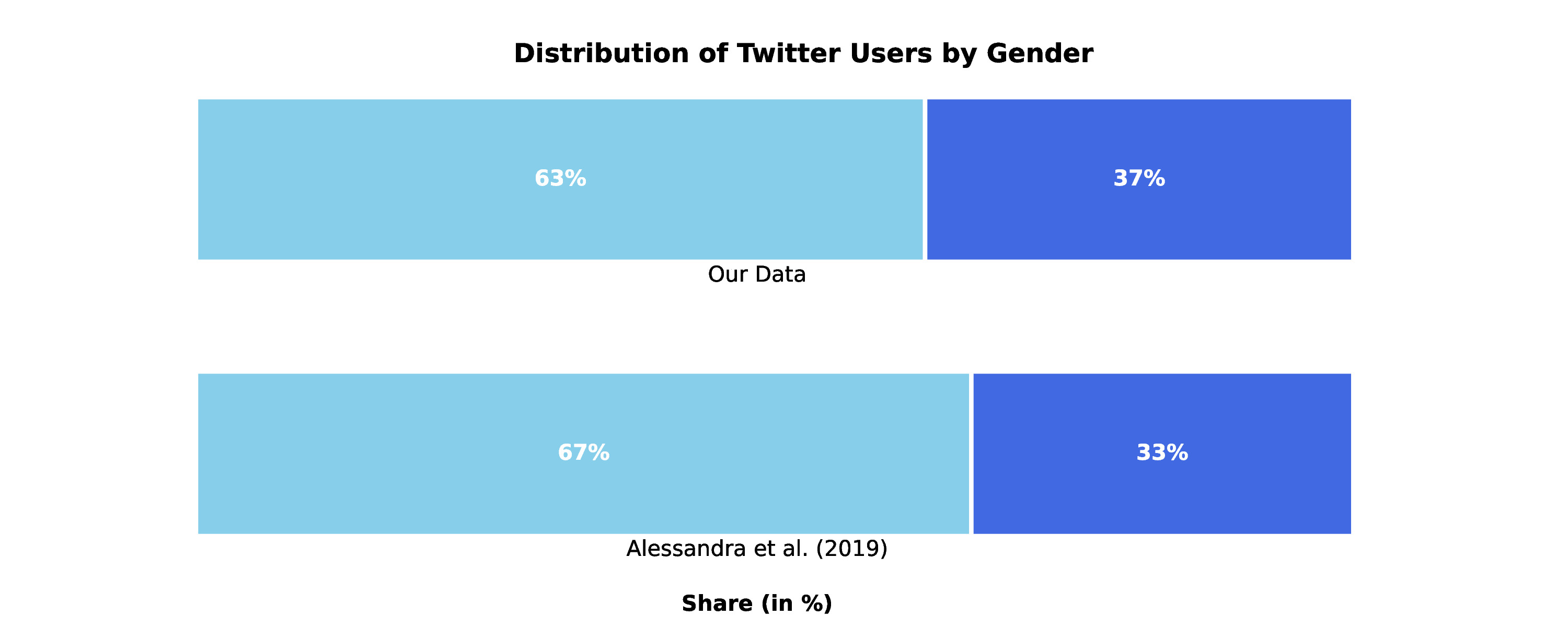}
    \end{subfigure} \hfill
    \begin{subfigure}{\linewidth}
        \centering
        \includegraphics[width=0.9\linewidth]{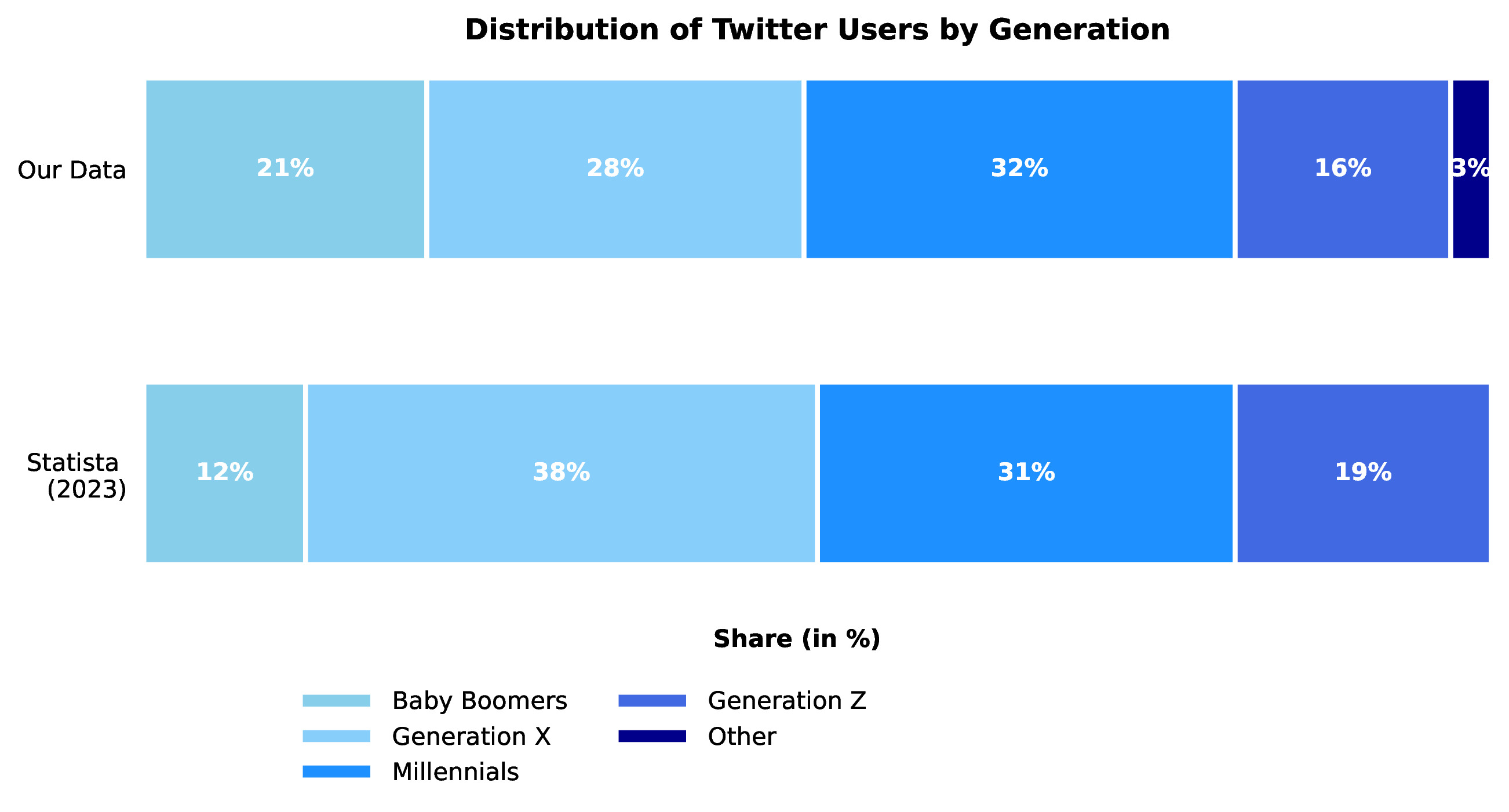}
    \end{subfigure}
    \caption{Representativness of Twitter users in DADIT.\label{fig:representativness}}
\end{figure*}

Overall, the demographic characteristics of users in our data align well with existing work (see \Cref{fig:representativness}). 63\% of our users are male relative to 67\% in the work by \cite{alessandra2019tweets}. Also, the age distribution of our users aligns closely with the reported age distributions of generations in \cite{statista_twitter}. The main difference is that we appear to observe more baby boomers among Twitter users in our data. This actually makes the age distribution of our Twitter users more representative of the age distribution of the Italian population overall. 

%%%%%%%%%%%%%%%%%%%%%%%%%%%%%%%%%%%%%%%%%%%%%%%%%%%%%%%%%%%%%
%%%%%%%%%%%%%%%%%%%%%%%%%%%%%%%%%%%%%%%%%%%%%%%%%%%%%%%%%%%%%

\begin{comment}
%%%%%%%%%%%%%%%%%%
\subsection{Additional Details on Data Annotation}
%%%%%%%%%%%%%%%%%%

\paragraph{Annotation of Gender}

\paragraph{Annotation of Age}

%%%%%%%%%%%%%%%%%%%%%%%%%%%%%%%%%%%%%%%%%%%%%%%%%%%%%%%%%%%%%
\begin{figure}[ht]
    \centering
    \includegraphics[width=0.4\textwidth]{example-image-a} 
    \caption{F1-Score fine-tuned models (tweets) \label{fig:model_performance_nr_tweets}}  
\end{figure}
%%%%%%%%%%%%%%%%%%%%%%%%%%%%%%%%%%%%%%%%%%%%%%%%%%%%%%%%%%%%%

\end{comment}

%%%%%%%%%%%%%%%%%%%%%%%%%%%%%%%%%%%%%%%%%%%%%%%%%%%%%%%%%%%%%%%%%%%%%%%%%%%%%%%%%%%%%%%%%%%%%%
%%%%%%%%%%%%%%%%%%%%%%%%%%%%%%%%%%%%%%%%%%%%%%%%%%%%%%%%%%%%%%%%%%%%%%%%%%%%%%%%%%%%%%%%%%%%%%
\section{Experimental Details \label{sec:appendix_exp}}
%%%%%%%%%%%%%%%%%%%%%%%%%%%%%%%%%%%%%%%%%%%%%%%%%%%%%%%%%%%%%%%%%%%%%%%%%%%%%%%%%%%%%%%%%%%%%%
%%%%%%%%%%%%%%%%%%%%%%%%%%%%%%%%%%%%%%%%%%%%%%%%%%%%%%%%%%%%%%%%%%%%%%%%%%%%%%%%%%%%%%%%%%%%%%
\begin{table}[ht]
    \centering
      \footnotesize
    \begin{tabular}{lc}
    \toprule
    \textbf{Model} &\textbf{\# Tokens} \\
    \midrule
M3                      & na\\
XLM                      & 512\\
Flan-T5                      & 512\\
GPT3.5                      & 2048\\
\bottomrule
\end{tabular} 
\caption{Models' max input lengths used in the experiments.\label{tab:prom}}
\end{table}

\begin{table*}[ht]
\centering
\footnotesize
\begin{tabular}{p{0.9\linewidth}}
    \toprule
    \textbf{Flan-T5}\\
    \midrule
    \texttt{Classify the gender (male or female) of the author of these texts:}\\
    \texttt{Bio: [input bio] Tweets: [input tweets]}\\
    \texttt{Gender:}\\

    \noalign{\vskip 2mm}
    \hdashline
    \noalign{\vskip 2mm}
    
    \texttt{Classify the age of the author of the following texts across 4 age groups: "0-19", "20-29", "30-39", "40-100". Texts:}\\
    \texttt{Bio: [input bio] Tweets: [input tweets]}\\
    \texttt{Age group:}\\
    \midrule
    \textbf{GPT3.5}\\
    \midrule
    \texttt{Classify the gender of the author of these texts (only answer with the label - "male" or "female" - even if uncertain!):}\\
    \texttt{[few shot examples]}\\
    \texttt{Bio: [input bio] Tweets: [input tweets]}\\
    \texttt{Gender:}\\
    \noalign{\vskip 2mm}
    \hdashline
    \noalign{\vskip 2mm}
    \texttt{Guess the age of the author of the following texts across 4 age groups: "0-19", "20-29", "30-39", "40-100". Only answer with an age group, even if uncertain.}\\
    \texttt{[few shot examples]}\\
    \texttt{Bio: [input bio] Tweets: [input tweets]}\\
    \texttt{Age group:}\\
    \bottomrule
\end{tabular} 
\caption{Language models' prompts for gender and age classification. Flan-T5 was prompted in a zero-shot setting, GPT3.5 in a few-shot settings (8 exemplars).\label{tab:prompts}}
\end{table*}

\begin{table}[ht]
    \centering
      \footnotesize
    \begin{tabular}{cccc}
    \toprule
    \textbf{\# tokens} & \textbf{\# shots} & \textbf{F1} & \textbf{Acc} \\
    \midrule
    512 & 0 & 75.2 & 76.1 \\
    1028 & 0 & 76.4 & 77.2 \\
    2056 & 0 & 76.9 & 77.7 \\
    2056 & 8 & 77.4 & 78.0 \\
    \bottomrule
    \end{tabular} 
\caption{GPT3.5 performance on Italian gender prediction with varying context lengths (\# tokens) and exemplars (\# shots). The longer the context, the more tweets from the user are included. \label{tab:gpt-ablation}}
\end{table}

\paragraph{M3}%Mahyar
We employed the PyTorch implementation of the M3 model to infer the age and gender of Twitter users \newcite{wang2019demographic}. M3 uses names, usernames, biographies, and profile images, but it does not utilize tweet content. For our analysis, we specifically excluded usernames from the input, replacing both names and usernames with empty strings, as detailed in our discussion. An essential requirement for M3 is the language of the user's biography; for this, we utilized the \emph{langdetect} package to identify the bio's language within M3's supported range. If the detected language was unsupported by M3 (rear cases), Italian was chosen as the default language. The model outputs probabilities for different gender and age groups. Our approach was to consider the group with the highest probability as the definitive prediction made by the M3 model.

\paragraph{Computer vision model}
We apply a popular open-source computer vision classifier \cite{levi2015age} for inferring age and gender from users' profile pictures. In a first step, the classifier identifies faces in the images and then classifies each face in terms of gender and age. The original application of the model uses the following age categories: [0, 2],[4, 6], [8, 12], [15, 20], [25, 32], [38, 43], [48, 53], [60, 100]. We convert these to the classes used by M3 by aggregating predicted probabilities for appropriate classes. If the profile image shows no face, we assign equal probability to all classes and then use the majority class ("male" for gender, "above 40" for age) as default. Finally, if there are multiple faces on the image, we assign the average predicted age and gender to the user.

\paragraph{XLM} 
For our classification task, we utilized the \texttt{twitter-XLM-roberta-base} model \cite{barbieri2021xlm}, a multilingual language model specifically fine-tuned on Twitter data. We trained two separate classification models for gender and age with the same fine-tuning settings for both models. Initially, we allocated 10\% of our training data as a validation set. The fine-tuning process employed the Adam optimizer, with an initial learning rate set at 2e-5 and a linear learning rate scheduler. The initial two epochs focused on warming up the gradients by training only the classification head. Following this, we unfroze the entire model for further training. We employed an early stopping mechanism, using F1 score as the criterion for improvement. Training was halted if there was no improvement in the F1 score on the validation set after three epochs. The model iteration that achieved the best F1 score on the validation set was retained as the final model.

\paragraph{Flan-T5} We experiment in a zero-shot setting with Flan-T5 because, in preliminary tests, it did not benefit from including exemplars in the prompt. When adding tweets as a feature, we do so until we saturate the maximum input length of 512 tokens. The prompts used are shown in Table \ref{tab:prompts}. Even if the input text is in Italian, prompts are in English as the model has been trained mostly with English instructions. Indeed, Italian prompts performed less well in our preliminary experiments for both Flan-T5 and GPT3.5. The generated output is parsed to find patterns corresponding to one of the labels. In case multiple labels are generated, we only consider the first one, while we assign the majority class label as the default in case nothing valid is generated. 

\paragraph{GPT3.5} In this case, we experiment in a few-shot setting, which proved to be significantly superior to zero-shot classification. When adding tweets as a feature, we do so until we saturate the maximum input length of 512 tokens. The prompts used are shown in Table \ref{tab:prompts}. The model benefits from longer context and few-shots, as shown in Table \ref{tab:gpt-ablation}. Although \texttt{gpt-3.5-turbo} has a maximum context length of 4096, we decided to run prompts with 2048 tokens maximum as a compromise between performance, monetary cost, and time. In fact, when adding tweets up to 4096 tokens, we needed to decrease the frequency of API requests so as not to exceed the maximum number of tokens allowed per minute. This would result in a significant slowdown of the classification process for a non-significant increase in performance.
As in the case of Flan-T5, the generated output is parsed to find patterns corresponding to one of the labels. In case multiple labels are generated, we only consider the first one, while we assign the majority class label as the default in case nothing valid is generated. 

When adding tweets as a feature, we do so until we saturate the maximum input length of 512 tokens. The generated output is parsed to find patterns corresponding to one of the labels. In case multiple labels are generated, we only consider the first one, while we assign the majority class label as the default in case nothing valid is generated. 

\end{document}